\documentclass[conference]{IEEEtran}
\IEEEoverridecommandlockouts
\usepackage{cite}
\usepackage{amsmath,amssymb,amsfonts}
\usepackage{algorithmic}
\usepackage{graphicx}
\usepackage{textcomp}
\usepackage{url}
\usepackage{xcolor}
\usepackage{multirow}
\usepackage[whole]{bxcjkjatype}
\def\BibTeX{{\rm B\kern-.05em{\sc i\kern-.025em b}\kern-.08em
    T\kern-.1667em\lower.7ex\hbox{E}\kern-.125emX}}
\begin{document}

\title{Indexing and Visualization of Climate Change Narratives Using BERT and Causal Extraction}

\author{\IEEEauthorblockN{Hiroki Sakaji}
\IEEEauthorblockA{\textit{Research Faculty of Information Science and Technology} \\
\textit{Hokkaido University}\\
sakaji@ist.hokudai.ac.jp}
\and
\IEEEauthorblockN{Noriyasu Kaneda}
\IEEEauthorblockA{\textit{Bank of Japan} \\
noriyasu.kaneda@boj.or.jp}
}

\maketitle

\begin{abstract}
\renewcommand{\thefootnote}{\fnsymbol{footnote}}
\footnote[0]{© 2023 IEEE.  Personal use of this material is permitted.  Permission from IEEE must be obtained for all other uses, in any current or future media, including reprinting/republishing this material for advertising or promotional purposes, creating new collective works, for resale or redistribution to servers or lists, or reuse of any copyrighted component of this work in other works.}
\renewcommand{\thefootnote}{\arabic{footnote}}
In this study, we propose a methodology to extract, index, and visualize ``climate change narratives'' (stories about the connection between causal and consequential events related to climate change). We use two natural language processing methods, BERT (Bidirectional Encoder Representations from Transformers) and causal extraction, to textually analyze newspaper articles on climate change to extract ``climate change narratives.'' The novelty of the methodology could extract and quantify the causal relationships assumed by the newspaper's writers. Looking at the extracted climate change narratives over time, we find that since 2018, an increasing number of narratives suggest the impact of the development of climate change policy discussion and the implementation of climate change-related policies on corporate behaviors, macroeconomics, and price dynamics. We also observed the recent emergence of narratives focusing on the linkages between climate change-related policies and monetary policy. Furthermore, there is a growing awareness of the negative impacts of natural disasters (e.g., abnormal weather and severe floods) related to climate change on economic activities, and this issue might be perceived as a new challenge for companies and governments. The methodology of this study is expected to be applied to a wide range of fields, as it can analyze causal relationships among various economic topics, including analysis of inflation expectation or monetary policy communication strategy.
\end{abstract}


\section{Introduction}
Climate change can potentially affect the economy and financial market through various pathways (it is called climate-related risk in economics), and addressing are being actively discussed domestically and internationally as a global issue. In response to the growing awareness of climate-related risk, various economic agents, including companies, households, financial institutions, and governments, have been progressing with climate change, which has begun to affect their economic activities and investment behaviors. In addition, research on the impact of climate change on the economy, market, and financial system is gaining momentum in finance and economics.

Understanding how economic entities have responded to the policy debate over climate change and the strengthening of the implementation of environmental regulations (e.g., CO2 emission reductions) related to climate change could help understand the direct and indirect impacts of climate change on the economy and consider appropriate policy responses.

In recent years, one approach to analyzing climate change issues has been to conduct text analysis of climate change news in an attempt to determine the level of interest and concern about climate change among the market and economy. For example, a study has been reported that constructs an index of the frequency of occurrence of news about climate change, which indicates the magnitude of concern about climate-related risk, and analyzes the relationship between stock returns and the news index's variation.

Text analysis can extract new information that cannot be ascertained from existing structural data. However, most text analysis technique used so far in the financial and economic fields is old-fashioned, which isolates text into word units, determining topics from the frequency of occurrence of certain words and co-occurrence relations between different words. They could not consider the context or logical structure in the text for understanding economic behaviors.

This study attempts to obtain implications regarding the path and magnitude of the impact of climate change issues on finance and the economy by conducting a syntactic analysis of the entire text without dropping information on the context and logical structure in the text. For example, suppose that in a text of corporate behavior on climate change as below.

``Companies have increased their green energy investments in response to future strict environmental regulations introduced by the authorities.``

Based on a variety of information above, including economic knowledge, IR information by companies, and industry outlook by experts, the writer of the text assumes a story in which the effect of ``increase in green energy investment in the industry'' is caused by the cause of "more strict environmental regulations by the authorities. In this study, we refer to the cause and effect relationship that constitutes the story assumed by the writer of the text as the causal relationship.

Analysis that takes into account causal relationships in the text can lead to a better understanding of the information behind economic events that cannot be understood from economic figures, i.e., how causal relationships are perceived, which may be beneficial for financial and economic forecasting as well as for testing economic and financial theories. Professor Robert Shiller of Yale University, Nobel laureate in economics, has advocated ``Narrative Economics'' which examines economic trends by analyzing the spread of narratives that express people's interests and emotions and how these narratives change over time (Shiller \cite{shiller2017narrative}). In addition, Shiller \cite{shiller2020narrative} presented the hypothesis that \textit{``An economic narrative is a contagious story that can change how people make economic decisions.''}

Based on this hypothesis, the inference is that narratives, when disseminated among many people, cause changes in people's expectations and behavior toward the future, resulting in changes in the macroeconomy. If we can extract the narratives that drive the economy from text data that many people read and see, such as literature, mass media, and social media, it would be possible to decipher the causal relationships behind the perceptions and actions of economic agents. This study refers to the extracted causal relationships in economic stories as economic narratives.

We propose a methodology for extracting, indexing, and visualizing economic narratives about climate change (referred to in this study as ``Climate change narratives'') by focusing on causal expressions assumed by the writer of a newspaper article using two natural language processing methods. The first is BERT (Bidirectional Encoder Representations from Transformers), a deep learning-based generic language model. The second is causal extraction, proposed by Sakaji et al. \cite{10.1007/978-3-540-89447-6_12}. Causal extraction is a method that searches for sentences containing causal relations in a text and extracts expression pairs that represent logical relations between cause and effect. Combining the results of the two methods, we can find causal relationships between cause and effect events across time points and topics, which can be extracted as ``Climate change narratives.'' The novelty of this study is that it analyzes contextual information in the text and extracts combinations of expressions in cause and effect relationships, thereby conducting text analysis that considers causal relationships.
The climate change narratives extracted in this study were indexed and visualized as a network diagram, suggesting the following four points.

(1) Climate change narratives that show causality to domestic environmental and energy policies tend to increase when events occur, such as a multilateral agreement at a high-level international climate change conference.

(2) Since 2018, there has been a marked increase in climate change narratives that focus on the causal link from domestic environmental and energy policies to corporate behavior and financial markets. This suggests that firms and markets have begun to change their perceptions and behaviors in anticipation of future environmental regulations.

(3) A new climate change narrative is emerging between ``international conferences and domestic environmental and energy policies'' and "monetary policy, prices, and business confidence.'' It appears that there is an awareness of the possibility that domestic and international climate change-related regulations may have some impact on macroeconomic activity. Narratives linking to the possibility of policy responses by central banks are also emerging.

(4) Climate change narratives related to natural disasters have increased recently. It is suggested that the adverse impacts of more severe and frequent natural disasters on economic activities have become a new risk factor for companies and authorities.

\section{Previous work}
This section introduces climate change news analysis using text analysis and economic and finance research using BERT and outlines how to find causal relationships in text.

\subsection{Climate Change News Analysis}
In recent years, many studies have analyzed climate change-related texts (news reports, newspaper articles, etc.) as one of the analytical approaches to climate change. For example, finance studies have used topic models such as Latent Dirichlet Allocation (LDA) to create climate change news indices and demonstrate the relationship with stock returns (Bua et al. \cite{bua2022}, Engle et al. \cite{engle2020hedging}, Faccini et al. \cite{faccini2023dissecting}, Hiraki et al. \cite{hiraki2023}, Pastor et al. \cite{pastor2022dissecting}). Fueki et al. \cite{fueki2022} emitters tend to curtail capital investment, especially firms. Arseneau et al. \cite{Arseneau2022} also analyzed central bank climate change-related speeches and found increasing references to the potential impact of climate change on macroeconomic, monetary policy, and financial stability. Thus, an increasing number of studies have used text analysis of climate change news to quantify the interest in and concern about climate change risks by markets, companies, and central banks, as well as to demonstrate the implications for finance and the economy.

\subsection{Economic and Finance Research Using BERT}
BERT is a deep learning-based general-purpose language model proposed by Devlin et al. \cite{Devlin2018}, which is used to extract important information from text and determine a sentence's topic.  BERT is a general-purpose language model based on deep learning that can be used to extract important information from text and determine a sentence's topic. BERT is published in a pre-trained general-purpose language (general grammar and vocabulary) using much text data. Therefore, users can obtain a training model for a specific task by additional learning (fine-tuning) using a small amount of labeled teacher data. The recent popularity of BERT is due to its versatility in enabling additional learning for different purposes.

In recent years, economic and finance research has begun to see studies that use BERT to analyze large text data, such as corporate financial statements and patent information, to perform text classification, information extraction, and financial data substantiation. In climate change-related analysis, Bingler et al. \cite{bingler2022cheap} extracted information on the progress of companies in dealing with climate change and evaluated their scores; Kölbel et al. \cite{kolbel2020ask} created a sentiment index for physical and transition risks and conducted empirical research on the relationship with CDS spreads for individual companies. Thus, previous studies have used large-volume data to build classification models of climate change-related financial and economic topics, created news indicators and score ratings in texts, and analyzed stock returns, corporate behavior, and economic activity.

\subsection{Causality in the Text}
There is not always a clear definition of causality, as described in the text. One linguistic explanation is that causal relations are included in sentences in which ``causes or reasons'' and ``facts that have already occurred'' are connected proximally (sentences in which logical sentences of cause and effect are connected by cue expressions such as ``because,'' ``because of,'' and ``due to,'' Iori \cite{iroi2012}).

Based on this view, this study uses ``causal extraction,'' a method proposed by Sakaji et al. \cite{10.1007/978-3-540-89447-6_12}, to find causal relations contained in texts. Causal extraction uses cue expressions to search for sentences in a text containing causal relationships and can extract expression pairs representing logical relationships between cause and effect. For example, the sentence "The subprime loan problem will cause a global economic recession" describes a causal relationship with ``subprime loan problem'' as the cause clause and ``global economic recession'' as the effect clause. In this way, causal relations in a sentence can be found by performing a dependency analysis based on cue expressions such as ``because of'' and ``due to.''

It should be noted that none of the previous studies analyzed indices based on the frequency of occurrence of words or articles, and none of them explicitly considered economic narratives contained within the text. A new contribution of this study is that we extracted causal narratives by conducting a text analysis considering causal relationships between causal and consequential events contained in news texts.


\section{How to Estimate Climate Change Narratives}
\begin{figure}[h]
 \begin{center}
 \includegraphics[width=\hsize]{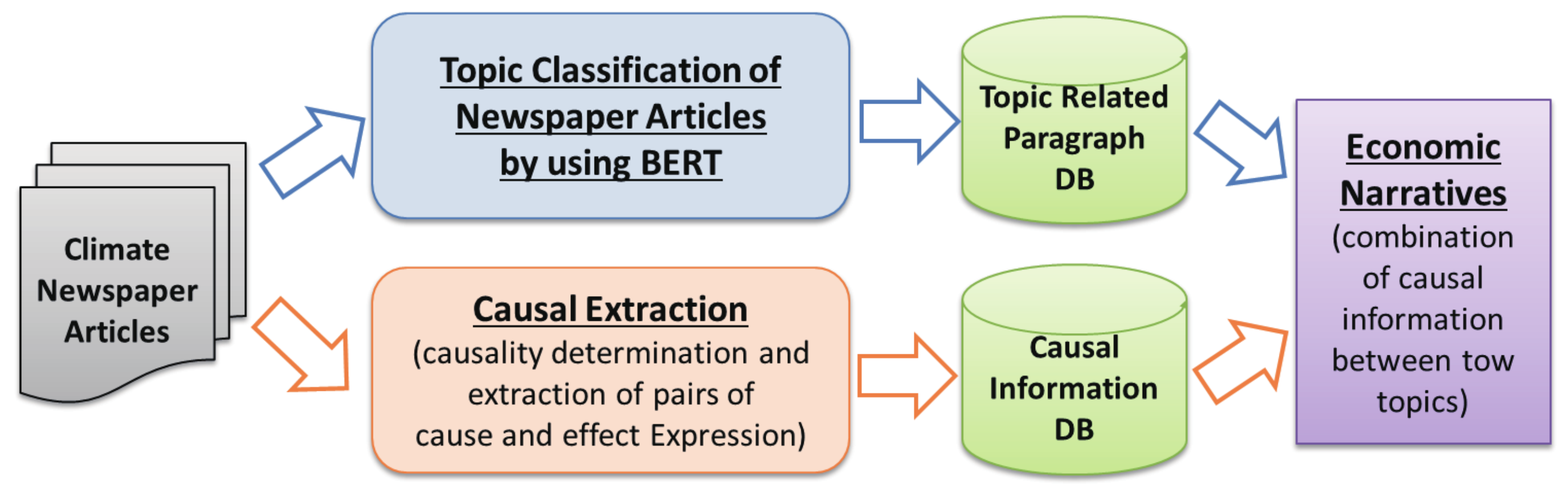}
 \end{center}
 \caption{Methodology for Extracting Climate Change Narratives}
 \label{img:overview1}
\end{figure}

\begin{figure}[h]
 \begin{center}
 \includegraphics[width=\hsize]{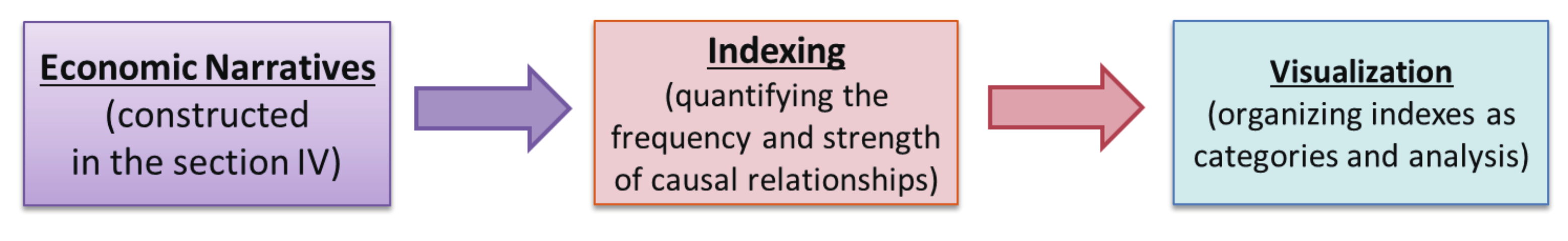}
 \end{center}
 \caption{Methodology for Indexing and Visualization of Climate Change Narratives}
 \label{img:overview2}
\end{figure}

The overall methodology for estimating climate change narratives proposed in this study is shown in Figure \ref{img:overview1} and Figure \ref{img:overview2} in this section. It consists of two parts: ``Methodology for Extracting Climate Change Narratives'' in section \ref{sec:ex_narrative} and ``Indexing and Visualization Methodology for Climate Change Narratives'' in section \ref{sec:indexing_narrative}.

\section{Methodology for Extracting Climate Change Narratives}
\label{sec:ex_narrative}

This section describes the data used to extract climate change narratives and the extraction methodology. In the following, we describe specifically the methodology for extracting climate change narratives using BERT and causal extraction.

\subsection{Data}
\begin{figure*}[h]
 \begin{center}
 \includegraphics[width=15cm]{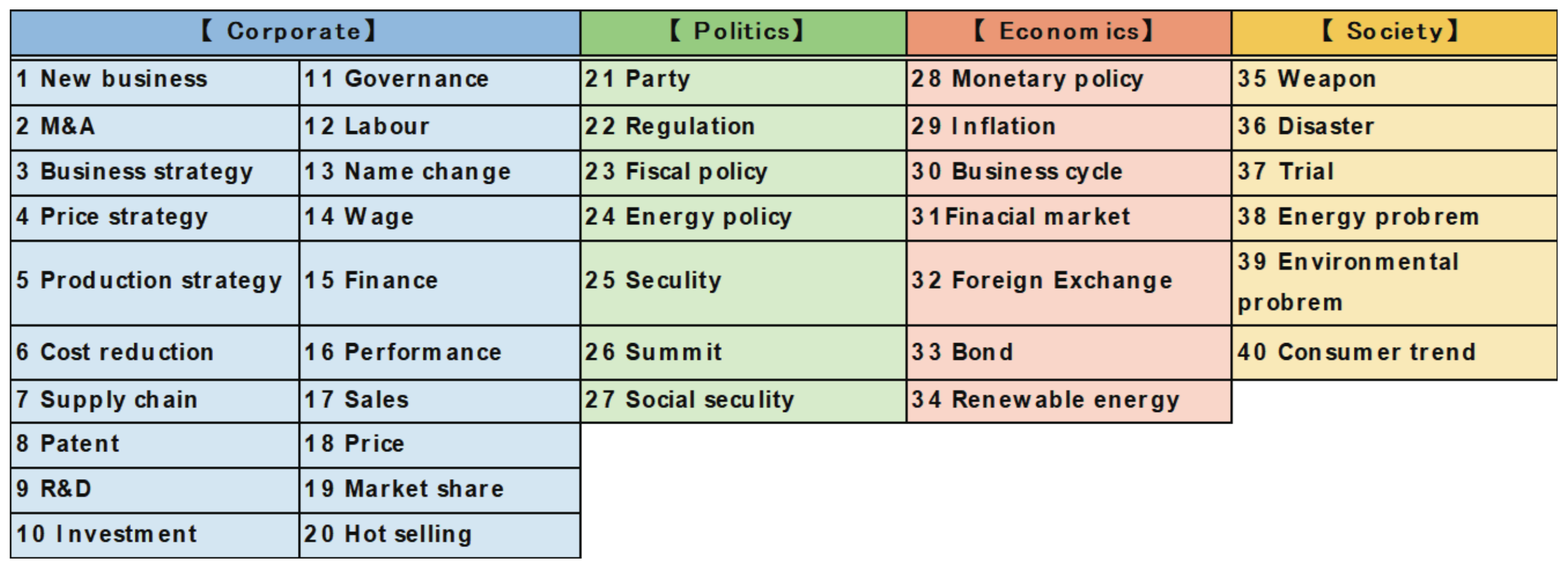}
 \end{center}
 \caption{Topic List}
 \label{img:topic_list}
\end{figure*}

This study analyzes a variety of climate change-related topics in \textit{the Nihon Keizai Shimbun} (an economic newspaper in Japan, morning edition and weekdays) from January 2000 to November 2021 (approximately 17,000 articles), including business-related, international conferences, domestic policy, and financial institutions. Each article is assigned a topic, and for this study, 40 topics related to corporate, politics, economics, and society were selected from topics set by the Nihon Keizai Shimbun (Figure \ref{img:topic_list}). These topics include those related to corporate behavior, macroeconomics, policy, regulation, politics, and society.

\subsection{Methodology for Extracting Climate Change Narratives}
This section describes the specifics of our methodology for extracting climate change narratives.
Since this study is dedicated to the analysis of finance-related documents in Japanese, we use ``Financial BERT'' (Suzuki et al. \cite{SUZUKI2023103194}), an additional pre-trained version of BERT (hereafter referred to as BERT).
First, classification accuracy was compared as a preliminary experiment to determine the methodology for topic classification of newspaper article paragraphs.
For each article in the Nikkei Shimbun, we checked the classification performance of various classification methods (BERT, linear regression, random forest, and support vector machine), assuming that the associated topics (classification word codes) were the teacher data (correct classification). For example, Table \ref{tb:olympic} shows a case study comparing the classification accuracy of articles related to the topic ``Olympics.''

\begin{table}[h]
  \caption{Classification results of the ``Olympic'' Classification Model.}
  \begin{center}
  \begin{tabular}{l|c|c|c}\hline
   & Precision & Recall & F1 \\ \hline \hline
   Linear Regression & 0.958 & 0.958 & 0.958 \\ \hline
   Random Forest & 0.951 & 0.950 & 0.950 \\ \hline
   SVM & 0.954 & 0.954 & 0.954 \\ \hline
   BERT & 0.960 & 0.960 & 0.960 \\ \hline
  \end{tabular}
  \end{center}
  \label{tb:olympic}
\end{table}

As a result of the comparison, the decision model by BERT shows a classification accuracy that slightly exceeds that of conventional machine learning methods, regardless of which criterion values are evaluated. In general, BERT is known to show high accuracy in various tasks, and based on the fact that it was confirmed that it could also perform with high accuracy in topic judgments, which is the subject of this study's analysis, BERT was employed in this study.

In the following, we propose a methodology for extracting climate change narratives that employs BERT. Specifically, it consists of STEP 1 through STEP 4.

\subsubsection{Step 1: Topic classification of newspaper articles by BERT}
BERT, which has trained a topic determination model on climate change, is used to determine the topics in newspaper articles (blue arrows in Figure \ref{img:overview1}). The specific procedure is as follows.

(1) Create teacher data for each topic using the taxonomy topic code (topic) assigned by default to the Nihon Keizai Shimbun articles. Some articles are assigned multiple taxonomy topic codes if they are written on multiple topics. When creating the teacher data, articles that are assigned a classifier code for the topic to be learned and that have no more than two assigned classifier codes are considered positive examples (correct answers), and articles that do not include the topic to be learned (only the classifier codes for other topics are assigned) are extracted as negative examples (incorrect answers).   

(2)The model is trained on each of the 40 topics to obtain each classification model for 40 topics.

(3) Using each classification model constructed in (2) above, determine whether or not each paragraph of all climate change-related newspaper articles is related to each of the 40 topics and record the classification results in the related paragraph database (DB). For each paragraph of all articles, 40 judgment flags are set for whether or not it is related to each of the 40 topics. By counting the number of paragraphs determined to be relevant to each topic, a ``climate change topic-based index'' can be created, similar to existing indexes in previous studies. By aggregating this index, it is possible to conduct a basic analysis of overall trends in climate change news (see section \ref{sec:indexing_narrative}).

\subsubsection{Step 2: Causal information is extracted from newspaper articles by causal extraction.}
Using the algorithm proposed by Sakaji et al. \cite{8285265}, we determine whether sentences containing causal relations exist in newspaper articles. Next, using causal extraction proposed by Sakaji et al. \cite{10.1007/978-3-540-89447-6_12}, expression pairs representing logical relationships between cause and effect are extracted from newspaper articles determined to contain causality and recorded as causal information (pairs of cause and effect expressions) in the causal information DB (orange arrow in Figure \ref{img:overview1}).

\subsubsection{Step 3: Extracting climate change narratives}
\begin{figure}[h]
 \begin{center}
 \includegraphics[width=\hsize]{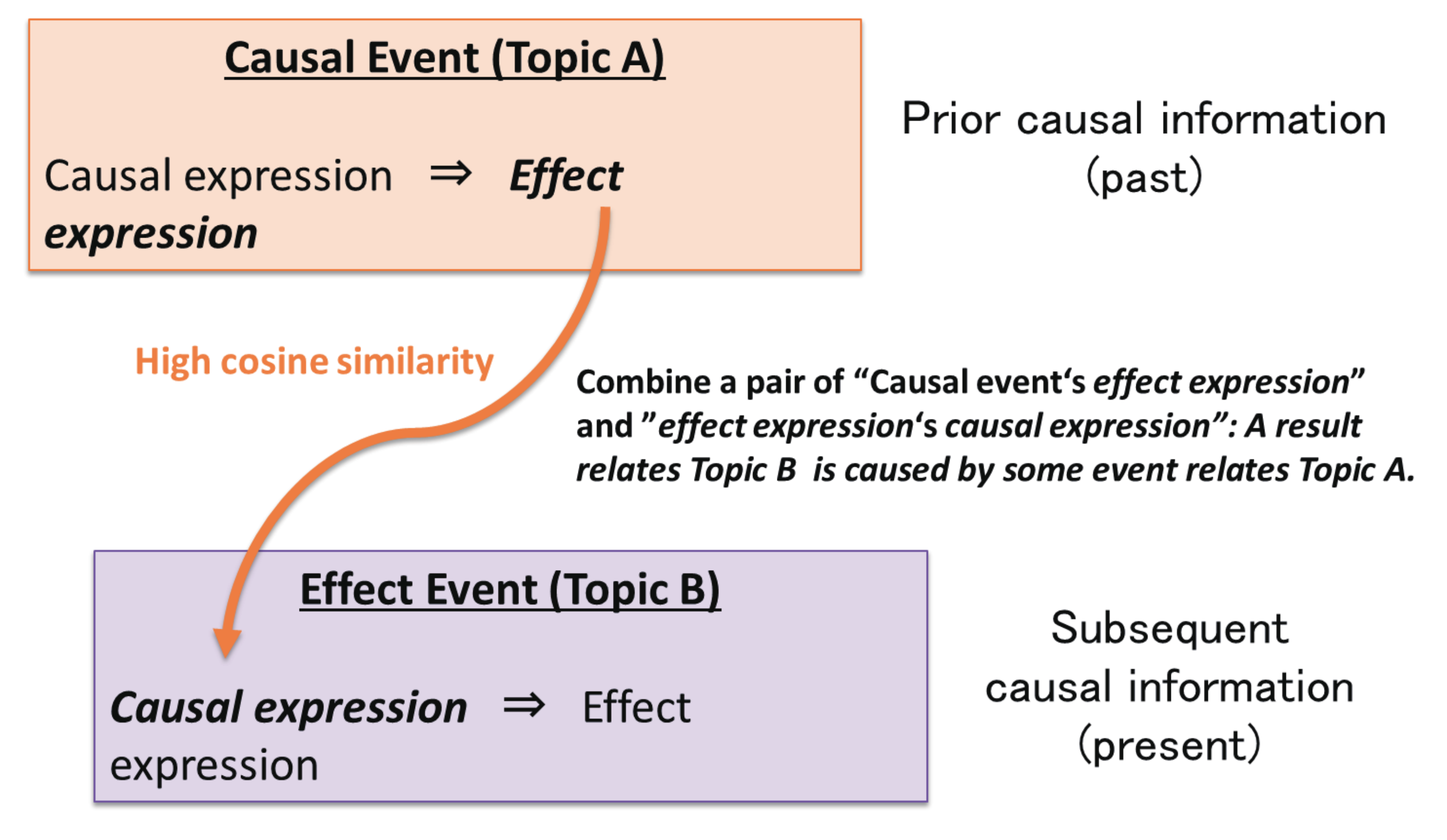}
 \end{center}
 \caption{Connecting Causals}
 \label{img:connect_causal}
\end{figure}

A pair of causal information across topics can be organized by linking the results of the topic determination in Step 1 and the causal information in Step 2. By pulling out the two types of information stored in the topic-related paragraph DB and the causal information DB and combining them as causal information in any two topics, we extract ``climate change narratives between two topics at different points in time'' (purple box on the left of Figure \ref{img:overview1}). In other words, we combine past causal events with current consequential events, assuming the temporal relationship that past news will spill over into future news. This work corresponds to the construction of causal chains proposed by Izumi and Sakaji \cite{10.1007/978-3-030-56150-5_2}. Below is an example of the coupling method when the causal event is an environmental regulation and the consequential event is a corporate strategy (Figure \ref{img:connect_causal}).

First, for each combination (pair) of a causal expression in a corporate strategy at the current time (result event) and a result expression in environmental regulation at the past time (cause event), its cosine similarity  (an indicator of how similar the expressions are) is calculated. If the cosine similarity of each pair exceeds a certain threshold, they are combined. Finally, a ``narrative is considered to exist'' between the topic to which the causal event belongs and the topic to which the consequent event belongs in the combined pair. That is as below,

\begin{itemize}
    \item A causality ``X then Y'' exists in an article on topic A.
    \item A causality ``Y' then Z'' exists in another topic B article at a point in time after the topic A article's publication date.
    \item ``Y and Y' have a high degree of similarity.''
\end{itemize}

By the transitional law, if three conditions are met, (i) ``Z if X'' holds, (ii) There exists a ``narrative'' (a connection between the causal and consequential events) from topic A, to which causal event X belongs, to topic B, to which effect event Z belongs.

\section{Methodology for Indexing and Visualization of Climate Change Narratives}
\label{sec:indexing_narrative}

The procedures described in the previous sections provide daily climate change narratives from one topic to another. In previous studies, these narrative indices can be applied to empirical studies using financial and economic data and climate change news indices.

In this section, first, time series data for the climate change topic-based index were generated using the same method as in previous studies to evaluate whether the classification of topics by BERT is valid. The same data are checked by comparing them with the timing of relevant international events. Next, we describe the methodology for indexing and visualizing climate change narratives proposed in this study. Climate change narratives (40 x 39 = 1,560) between causal events (40 topics) and consequential events (39 topics) are extracted. The resulting climate change narrative indices are indexed (purple arrows in Figure \ref{img:overview1}), and the indices showing particularly important connections are selected and visualized as a network association diagram (red arrows in Figure \ref{img:overview1}). Finally, the results of the indexing and visualization of the climate change narratives are analyzed for economic interpretation.

\subsection{Indexing and Visualization Methodology}
\subsubsection{Climate Change Topic-based Index}
\begin{figure}[h]
 \begin{center}
 \includegraphics[width=\hsize]{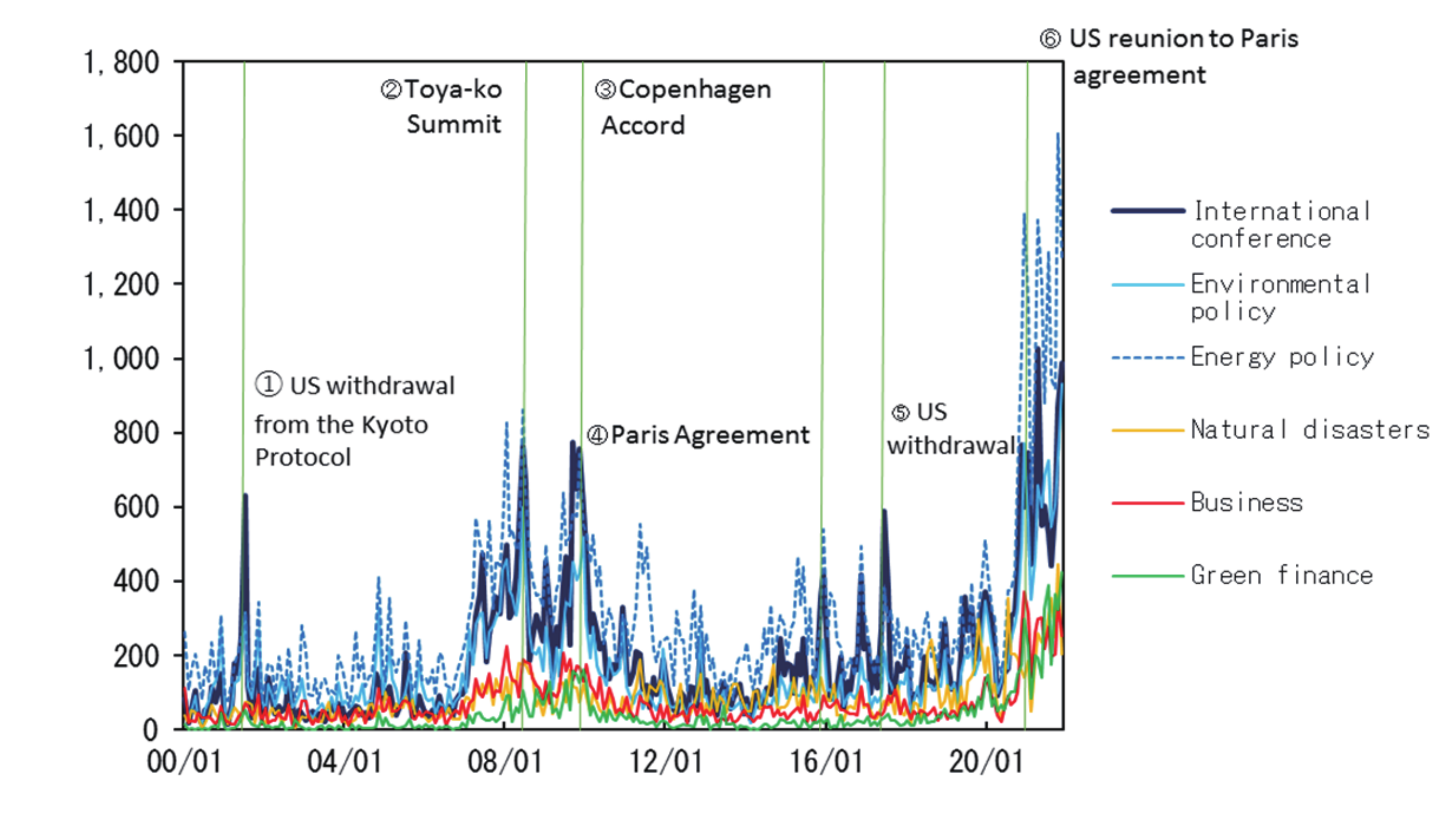}
 \end{center}
 \caption{Climate Change Index}
 \label{img:index}
\end{figure}

First, we calculate the ``climate change topic-based index,'' which is a monthly count of the number of paragraphs judged to be related to a given topic, using only the results of BERT judgments; as in previous studies using BERT, we index the number of occurrences of sentences containing a given topic. As shown in Figure \ref{img:index}, the index increases when events such as 2008 (G8 summit at Lake Toya), 2009 (Copenhagen Accord), and 2015 (Paris Agreement) occur, indicating a trend of increased coverage of climate change around the time of summits and international agreements. In addition, around 2008-2010, most topics were related to international discussions and domestic environmental regulations, reflecting that climate change was a central topic of domestic and international policy discussions.

On the other hand, especially since 2018, there has also been an increase in topics related to business strategies and markets, such as corporate climate change responses and coverage of green finance by financial institutions. This indicates that the coverage of climate change is spreading to various topics.

The index is useful in that it allows us to identify events that are important to the climate change debate, get a rough idea of the main topics at any given time, and correlations between topics. However, by its nature as an index that calculates the number of occurrences of relevant paragraphs by topic, it does not consider causality among topics.

\subsubsection{Climate Change Narrative Index}
Next, we explain how to create a climate change narrative index that considers causal relationships between topics. We index the ``two-topic climate change narratives between different points in time'' as in Equation \ref{eq:indexing}.

%
\begin{equation}
  Index\_monthly\_m = \sum^{M}_{j=0} \sum^{L(j)}_{i=0} \frac{1}{1+ae^{bd}} cos(\overrightarrow{i}_{t-d} \cdot \overrightarrow{j}_{t})
  \label{eq:indexing}
\end{equation}
\begin{equation}
  cos(\overrightarrow{i}_{t-d} \cdot \overrightarrow{j}_{t}) = \frac{\overrightarrow{i}_{t-d} \cdot \overrightarrow{j}_{t}}{|\overrightarrow{i}_{t-d}| |\overrightarrow{j}_{t}|}
  \label{eq:indexing}
\end{equation}

Here, 

\begin{description}
    \item[$M$ :] set of causal chains included in month $m$.
    \item[$L(j)$ :] set of cause event $\overrightarrow{i}_{t-d}$ connected to result event $\overrightarrow{j}_{t}$.
    \item[$t-d$ :] observation point of cause event leading to result event ($d > 0$).
    \item[$t$ :] observation point of result event included in month $m$.
    \item[$d$ :] time difference (days) between cause event and result event.
\end{description}

A monthly index is created by narrowing down and aggregating combinations with a high degree of cosine similarity ($cos(\overrightarrow{i}_{t-d} \cdot \overrightarrow{j}_{t}) \geqq 0.7$) in the causal chain of cause event vector $\overrightarrow{i}_{t-d}$ and result event vector $\overrightarrow{j}_{t}$.
Because information increases over time, a bias can arise that the more recent the news, the more connections to the past. We decrease the weight of older causal relationships over time to exclude this. Specifically, we assume that the logistic function is followed, and the parameters $a$ and $b$ are set according to the decay period of the news. In this study, the weights are set to decrease by half after five years.

\begin{figure}[h]
 \begin{center}
 \includegraphics[width=\hsize]{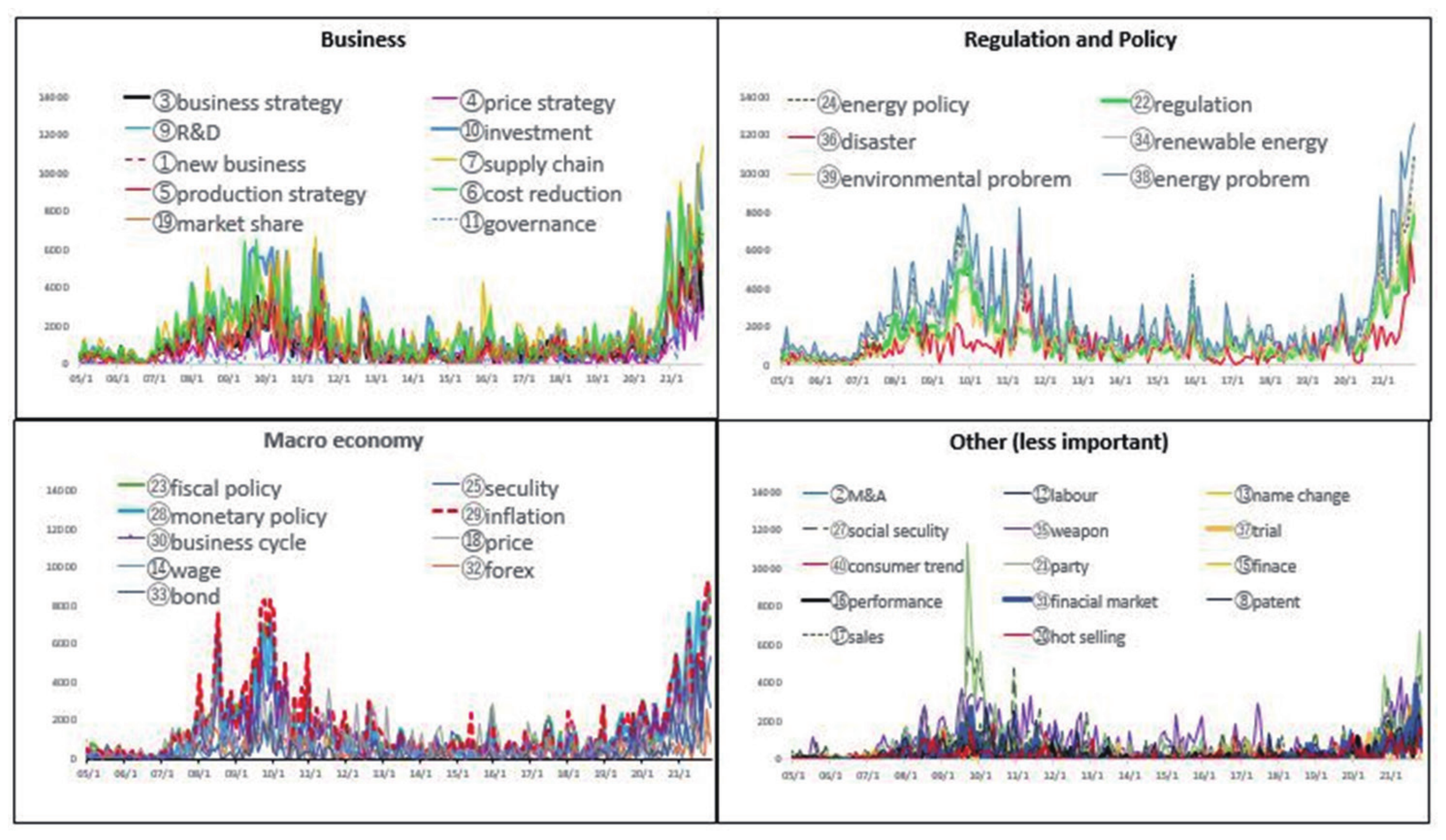}
 \end{center}
 \caption{Climate Change Narrative Index}
 \label{img:narrative_index}
\end{figure}

Next, we estimated 40 topics $\times$ 39 topics = 1,560 climate change narrative indices by specifying the order of the causal and consequential events, considering the causal relationship between any two topics. Figure \ref{img:narrative_index} illustrates the climate change narrative indexes from the causal event ``International Conference'' to the resultant events (the other 39 topics). It shows that the indices related to companies, regulations and institutions, and macro indicators and policies increased around 2008 to 2010 and have also increased significantly and reached high levels in recent years. The results also confirmed that the magnitude of the climate change narrative has changed from point in time to point in time. On the other hand, we find that there is little increase in topics that are less related to climate change. Thus, by comprehensively estimating various topics, we can let the data speak for itself, search for causal relationships, and examine the presence or absence of causal relationships related to climate change based on the trends in the index.

\subsubsection{Visualization Using Network-Related Diagrams}
A network association diagram visualizes how the topics that makeup climate change news are coupled with each other. Combining the topic-specific climate change narrative indexes reveals causal connections (one-way/mutual direction) between topics and provides an overall picture of the narratives within climate change-related news.

The Climate Change Narrative Index indicates the extent to which current news is influenced by past news; the stronger the influence, the higher the index. In this section, we focus on the period 2018-2021, when many climate change narrative indices rose significantly and new inter-topic narratives emerged. We find that the combinations of significantly rising climate change narrative indices are mainly related to the following categories: international conferences, domestic policy, business-related, and monetary policy/macroeconomics. We consider these to have been the central climate change topics in recent years and analyze them.

\begin{figure}[h]
 \begin{center}
 \includegraphics[width=\hsize]{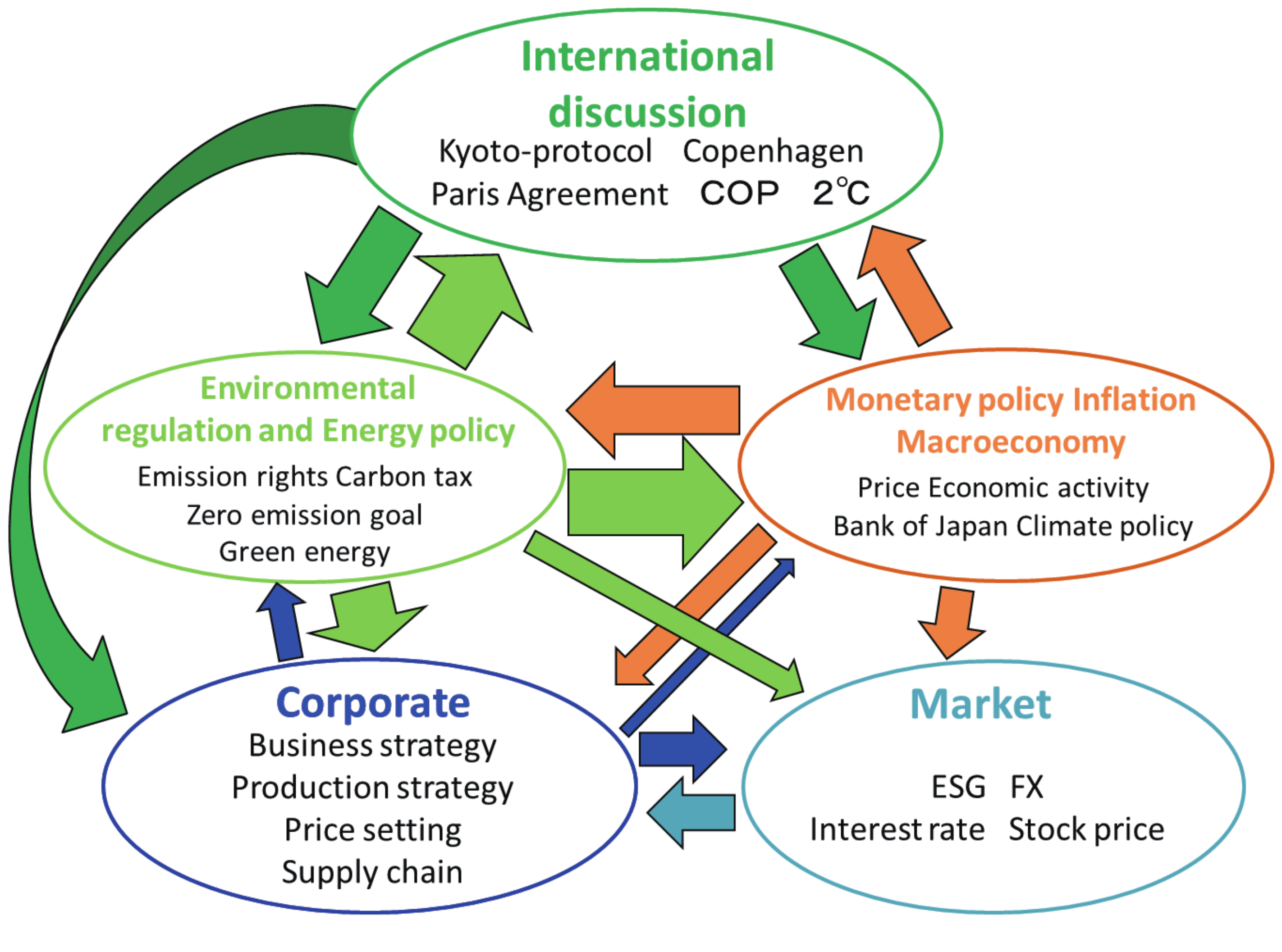}
 \end{center}
 \caption{Climate Change Narrative Graph}
 \label{img:narrative_graph}
\end{figure}

Figure \ref{img:narrative_graph} shows a network association diagram that displays the connections of the Climate Change Narrative Index by major category in aggregate. The size of the arrows connecting the categories roughly corresponds to the Climate Change Narrative Index level, meaning the strength of the causal relationship between the topics. In addition, for each category, a selection of the main keywords that appeared in the newspaper articles are listed.

\begin{figure}[h]
 \begin{center}
 \includegraphics[width=\hsize]{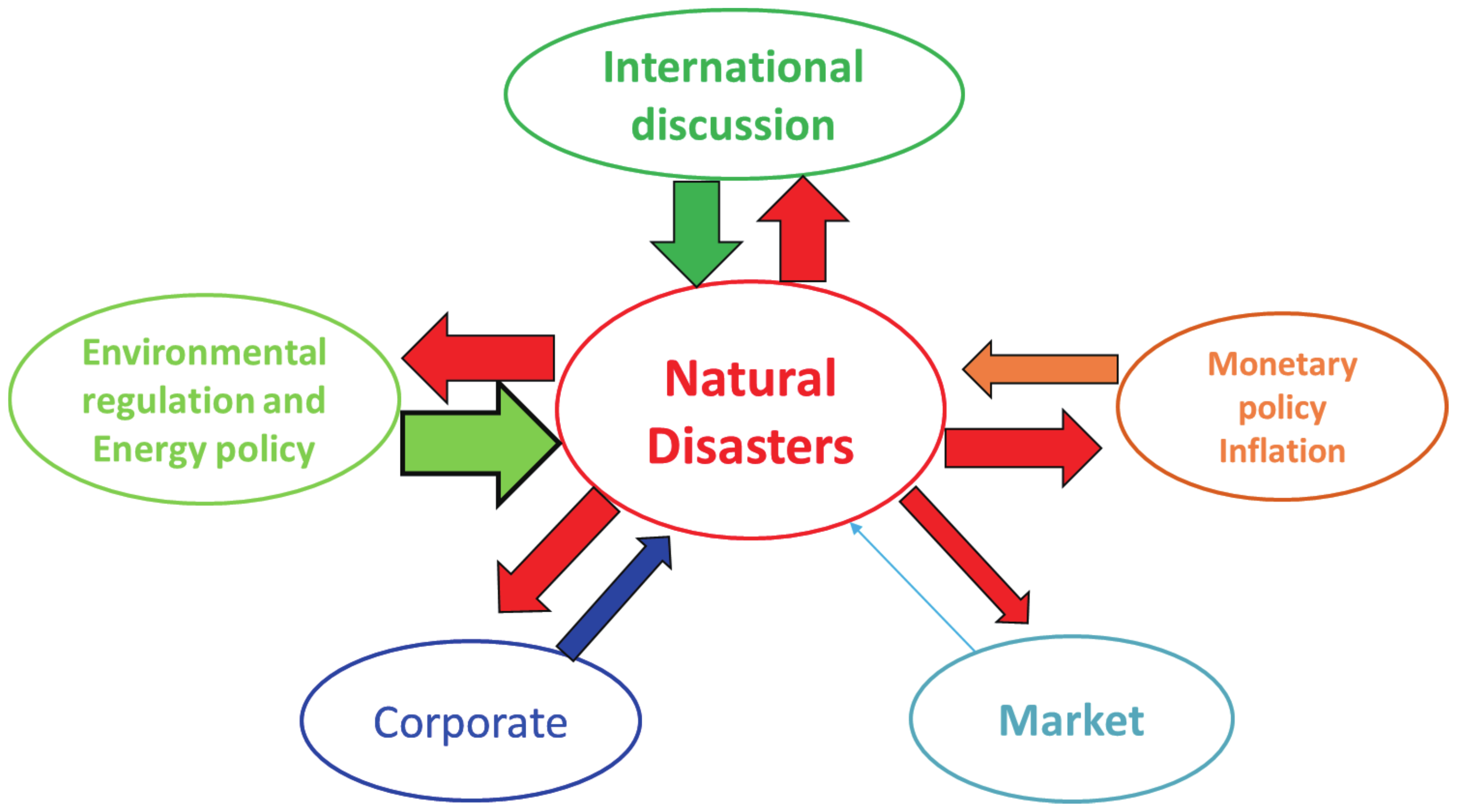}
 \end{center}
 \caption{Visualization of Economic Narrative for Natural Disasters
}
 \label{img:narrative_graph2}
\end{figure}

A network association diagram centered on natural disasters (e.g., flooding) that appear to be caused by climate change can also be created as a climate change narrative for a different pathway (Figure \ref{img:narrative_graph2}).

\section{Key Findings from Analysis of Climate Change Narratives}
The climate change narrative index and network diagram analysis examined in the previous section suggest the following four points.

(1) The time series data of the Climate Change Narrative Index (Figure \ref{img:narrative_index}) shows that the climate change narrative strengthens during events such as climate change-related international conferences and agreements. Examples include the G8 Hokkaido Toyako Summit in 2008, the Copenhagen Accord in 2009, the adoption of the Paris Agreement in 2015, and the US return to the Paris Agreement in 2021. Figure \ref{img:narrative_index} (top right) shows that climate change narratives, which show causality from international conferences (international climate change discussions) to regulatory and institutional-related (e.g., governments setting policy targets for decarbonization and green energy), have remained consistently higher than other topics, indicating that climate change has been a major news This suggests that climate change has been a major topic in the

(2) Since the ``2°C target'' was agreed to in the Paris Agreement in 2015, countries' climate change-related policies have shifted from the stage of broad, abstract policy discussions to the stage of considering policy implementation, such as strengthening environmental regulations. Looking at the middle part of Figures \ref{img:cross} and \ref{img:cross2} (rows 22-26 of the missing direction topic) and Figure 8, there has been a marked increase in climate change narratives for corporate-related issues (e.g., business strategy and capital investment) from strengthened domestic environmental and energy policies since 2018. Looking at the upper part of Figures \ref{img:cross} and \ref{img:cross2} (rows 1-10 of the missing direction topic), the corporate-related narratives for topics such as purchasing and procurement, capital investment, and business strategy have strengthened. This period suggests that companies and markets have begun to change their perceptions and behaviors in anticipation of future tighter environmental regulations and the associated concerns that transition risks will materialize.

(3) Looking at the bottom of Figures \ref{img:cross} and \ref{img:cross2} (lines 28-30 of the missing direction topic) and Figure \ref{img:narrative_graph}, a new inter-directional link has emerged in the climate change narrative between domestic and international environmental regulations and energy policies and monetary policy, prices, and business sentiment since 2018. It can be assumed that progress in international climate change-related policies and cross-industry climate change responses have become increasingly aware of the changes that macroeconomic activity and prices can cause. It will also likely reflect increasing talk about central bank policy responses to climate change.

(4) Figure \ref{img:narrative_graph2} suggests that climate change narratives related to natural disasters have strengthened in recent years in response to the intensification and increased frequency of floods and other natural disasters. In other words, it is suggested that climate change may be perceived as a new risk factor for corporate managers and authorities, as the negative impacts of climate change on economic activities through natural disasters are now recognized in addition to strengthening regulations. At the same time, the climate change narrative, from climate change-related policies to natural disasters, is increasing. Promoting stricter environmental regulations and green energy policies will likely mitigate future physical risks.

These are the results of analyzing the climate change narratives obtained in this study. The climate change narratives in this study allow us to examine the overall picture of climate change news by comparing time-series changes among various topics. When interpreting the results of the indices and network association diagrams, it is important to consider whether the movements are consistent with economic and finance theory.

\section{Summary and Discussion}
This study proposed a method to extract, index, and visualize ``climate change narratives'' using BERT and causal extraction. Specifically, we quantified "climate change narratives at different times and among different topics" and presented a method for indexing climate change narratives. Furthermore, we visualized the major climate change narratives as a network association diagram and presented a picture of causal connections surrounding climate change-related news.

Interpretation of the Climate Change Narrative Index suggested that the process of progress in international discussions on climate change and the concretization of domestic environmental regulations and energy policies has begun to change the perceptions and behaviors of companies, market participants, and others. It was also suggested that new links between domestic and international environmental regulations and macroeconomic, price, market, and monetary policies may emerge. The analytical results obtained using this methodology provide material for examining the impact of climate change on the macro economy, prices, and markets, as well as the possibility of policy responses by central banks, and will provide new insights for future data-based verification and other studies.

Finally, we discuss the causal relationships captured by the ``economic narratives'' in this study, not limited to the analysis of climate change news analyzed here, and discuss future issues.

As previously mentioned, Professor Shiller pointed out that ``An economic narrative is a contagious story that can change how people make economic decisions'' (Shiller \cite{shiller2020narrative}). It has been suggested that narratives that express people's interests and emotions are shared among many people through news (spreading throughout society), which may motivate individual economic activity and ultimately affect the macroeconomy. For example, Goetzmann, Kim, and Shiller \cite{goetzmann2022crash} extracted narratives about investors' crash fears (``crash narratives'') from the newspapers on Black Monday in 1987, the miserable U.S. stock market crash and found that the narratives significantly explained variation in the volatility of U.S. stocks.

This study proposes a new approach to extracting economic narratives from text data, indexes the strength of the narratives on the theme of climate change, and attempts to interpret the index. On the other hand, it is also possible to empirically analyze how the narrative index created in this study relates to financial and economic data, such as stock prices, and various ways of utilizing the narrative index.

In addition, statistical causal inference, which has been the focus of much attention in recent years, is an approach that statistically examines the causal relationships between causes and effects among data. Although the approach and methodology used in this study to search for economic narratives from the text are different, the two are considered complementary in verifying causal relationships; Shiller \cite{shiller2020narrative} notes that it is important to determine whether economic narratives indicate the true causes and consequences of economic events and notes that narrative analysis. For example, by integrating ``narratives'' contained in textual data and ``statistical information'' obtained by analyzing numerical data, it would be possible to elaborate further on the structure of causal relationships surrounding the financial economy through a different approach. For example, verifying whether the economic narratives extracted by this research approach have statistical causality (Granger causality, structural causality, etc.) is possible.

Other directions for future research development include applying our approach to financial and economic topics other than climate change (e.g., prices, economic fluctuations, etc.). For example, the ``economic narratives'' extracted from the textual data could be applied to forecast financial and economic data and test macroeconomic theories by identifying topics and concerns that firms and market participants have focused on during periods of economic uncertainty in the past. Specifically, the following problem settings could be considered: (1) to understand the topics that people were paying attention to in the context of the occurrence of bubbles (deviations from asset price fundamentals) and their collapse, and (2) to examine how narratives about price and wage trends shape household and firm inflation expectations. Analyzing how social norms (norms) are formed, for example, avoiding price hikes but not wage increases, is an important research topic.

Through the analysis of economic narratives, it is possible to examine the processes by which economic agents perceive phenomena and change their perceptions, as well as to analyze how they form expectations about the future. By clarifying the expectation formation mechanism, it will be possible to improve how policymakers disseminate information. Such applied research is considered to be an important research theme for the future.

\section*{Notes}
The opinions expressed in this article belong to the authors alone and do not represent the official views of the Bank of Japan. Additionally, all possible errors are solely the author's own.

\section*{Acknowledgments}
This work was supported in part by JSPS KAKENHI Grant Number JP21K12010 and JST-PRESTO Grant Number JPMJPR2267, Japan.

\bibliographystyle{IEEEtran}
\bibliography{list.bib}

\section*{Appendix}

\begin{figure*}[h]
 \begin{center}
 \includegraphics[width=\hsize]{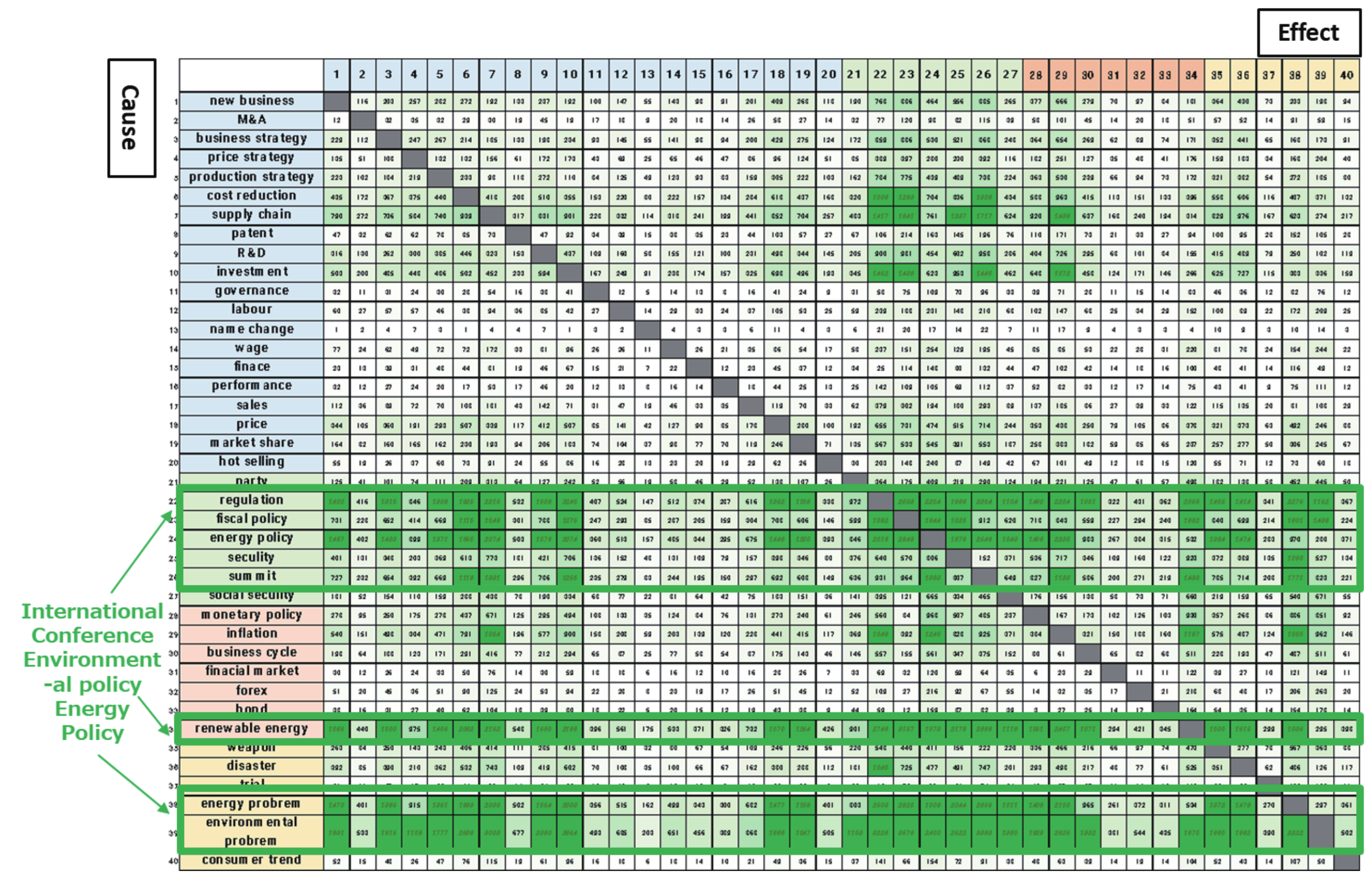}
 \end{center}
 \caption{Matrix of Climate Narratives (2000-2017 index average)}
 \label{img:cross}
\end{figure*}

\begin{figure*}[h]
 \begin{center}
 \includegraphics[width=\hsize]{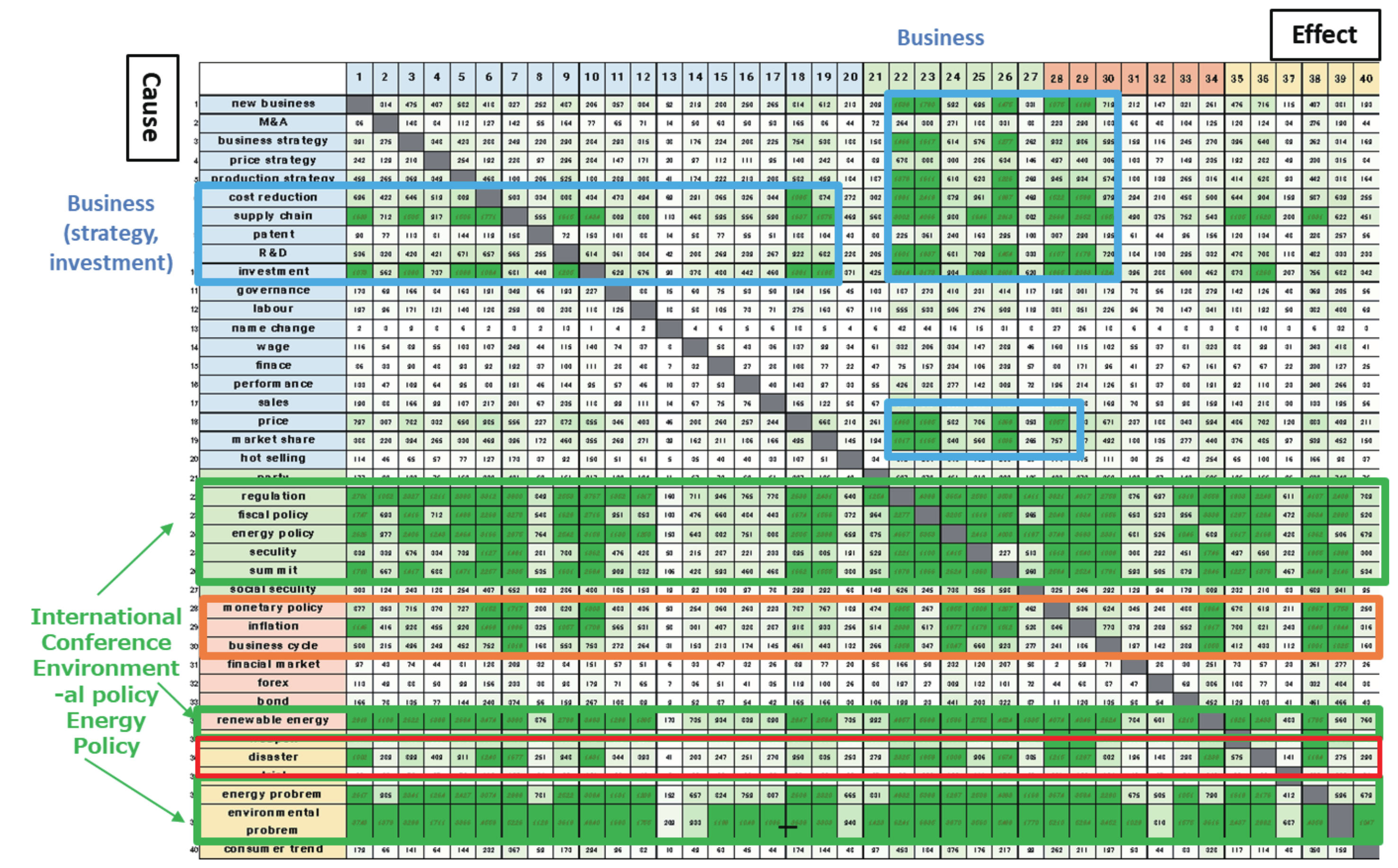}
 \end{center}
 \caption{Matrix of Climate Narratives (2018-21 index average)}
 \label{img:cross2}
\end{figure*}

Figure \ref{img:cross} calculates the monthly averages of the estimated 1,560 climate change narrative indices from 2000 to 2017 and displays them in a matrix (rows refer to the topics of causal events, and columns refer to the topics of consequential events). Figure \ref{img:cross2} similarly displays a matrix of monthly averages from 2018 to 2021. The higher the climate change narrative index level between each topic (stronger linkages), the darker green color is displayed.

Comparing Figures \ref{img:cross} and \ref{img:cross2}, the dark green areas are increasing, suggesting that climate change narratives are strengthening across various topics. In fact, looking at the movement from 2000 to 2017 (Figure \ref{img:cross}), it can be observed that narratives have only been gathered on topics related to international conferences on climate change and environmental and energy policy. On the other hand, looking at the movement from 2018 to 2021 (Figure \ref{img:cross2}), we can confirm that narratives continue to be concentrated on a wide range of topics from international conferences on climate change and environmental and energy policy. In addition, we can confirm that narratives have emerged between business-related issues and environmental and energy policies, within business-related issues (from capital investment and procurement to business strategy and price trends), and also between climate change-related regulations and natural disasters and monetary policy, prices, and business confidence. Thus, many climate change narrative indices have increased significantly in recent years, confirming the emergence of new inter-topic narratives.

\end{document}